\documentclass[final]{elsarticle}
\usepackage[a4paper]{geometry}
\usepackage[utf8]{inputenc}
\usepackage{graphicx}
\usepackage{amssymb}
\usepackage{color}
\usepackage{lineno}
\usepackage[ruled,vlined,linesnumbered,titlenumbered]{algorithm2e}
\usepackage{amsmath}
\usepackage{times}
\usepackage{optidef}
\usepackage{multirow}
\usepackage{makecell}
\usepackage{enumitem}
\usepackage{booktabs}
\usepackage[edges]{forest}
\usepackage{pifont}
\usepackage{soul}
\usepackage{amssymb}

\useforestlibrary{linguistics}
\forestapplylibrarydefaults{linguistics}

\journal{}

\begin{document}

\begin{frontmatter}


\title{Evolutionary Multitask Optimization: Fundamental Research \\ Questions, Practices, and Directions for the Future}

\author[tec]{Eneko Osaba\corref{cor1}}
\ead{eneko.osaba@tecnalia.com}
\author[tec,upv]{Javier Del Ser\corref{cor1}}
\ead{javier.delser@tecnalia.com}
\author[qtr,ntu]{and Ponnuthurai N. Suganthan}
\ead{p.n.suganthan@qu.edu.qa}

\address[tec]{TECNALIA, Basque Research and Technology Alliance (BRTA), P. Tecnologico, Ed. 700, 48160 Derio, Spain}
\address[upv]{University of the Basque Country (UPV/EHU), 48013 Bilbao, Spain}
\address[qtr]{KINDI Center for Computing Research, College of Engineering, Qatar University, Doha, Qatar}
\address[ntu]{School of Electrical \& Electronic Engineering, Nanyang Technological University, Singapore}
\cortext[cor1]{Corresponding authors. TECNALIA, Basque Research \& Technology Alliance (BRTA), P. Tecnologico, Ed. 700. 48170 Derio (Bizkaia), Spain. Both authors have contributed equally to the work.}

\begin{abstract}
Transfer Optimization has gained a remarkable attention from the Swarm and Evolutionary Computation community in the recent years. It is undeniable that the concepts underlying Transfer Optimization are formulated on solid grounds. However, evidences observed in recent contributions confirm that there are critical aspects that are not properly addressed to date. This short communication aims to engage the readership around a reflection on these issues, and to provide rationale why they remain unsolved. Specifically, we emphasize on three critical points of Evolutionary Multitasking Optimization: i) {\color{black}the plausibility and practical applicability of this paradigm}; ii) the novelty of some proposed multitasking methods; and iii) the methodologies used for evaluating newly proposed multitasking algorithms. As a result of this research, we conclude that some important efforts should be directed by the community in order to keep the future of this promising field on the right track. Our ultimate purpose is to unveil gaps in the current literature, so that prospective works can attempt to fix these gaps, avoiding to stumble on the same stones and eventually achieve valuable advances in the area.

\begin{changemargin}{1.0cm}{1.0cm} 
	DISCLAIMER: This paper has been accepted and published in Swarm and Evolutionary Computation under this DOI: https://doi.org/10.1016/j.swevo.2022.101203
\end{changemargin}

\end{abstract}


\end{frontmatter}

\section{Introduction} \label{sec:intro}

{\color{black}Addressing several learning tasks simultaneously by exploiting commonalities between them has been a central topic of research, ignited by an ever-growing substrate of practical contexts where multitasking can be framed. From the perspective of modeling problems, in the last decade} the existence of high-quality data has become a common factor in almost any discipline of knowledge. Industrial sectors traditionally reluctant to the advent of digital technologies ({\color{black}e.g., energy distribution or manufacturing plants}) have gone at a par with this vigorous information blossoming. As a result, Artificial Intelligence has encountered a magnificent opportunity to provide practical value and achieve unprecedented levels of performance over complex modeling tasks. 

In this context, the increase in the number of tasks and data flows that coexist in a certain scenario has motivated a major shift towards Artificial Intelligence algorithms capable of coping with several tasks. This is the case of multitask learning \cite{zhang2021survey}, which focuses on the development of learning models (e.g., image classification) that can address several related tasks at the same time; or continual learning \cite{delange2021continual}, {\color{black}which generalizes the former paradigm to handle modeling tasks that can dynamically emerge, change, or disappear over time.} Gains derived from addressing these tasks via multitask learning come in the form of shorter training periods, models of smaller size or a lower demand for annotated data, all benefiting from the proper exploitation of the synergistic commonalities between the modeling tasks at hand. Besides multitask learning, other research areas in Machine Learning that also promote the exchange of information between modeling tasks to boost their generalization performance are transfer learning \cite{zhuang2020comprehensive} and domain adaptation \cite{wang2018deep}.

\begin{figure}[h!]
	\centering
	\includegraphics[width=0.6\columnwidth]{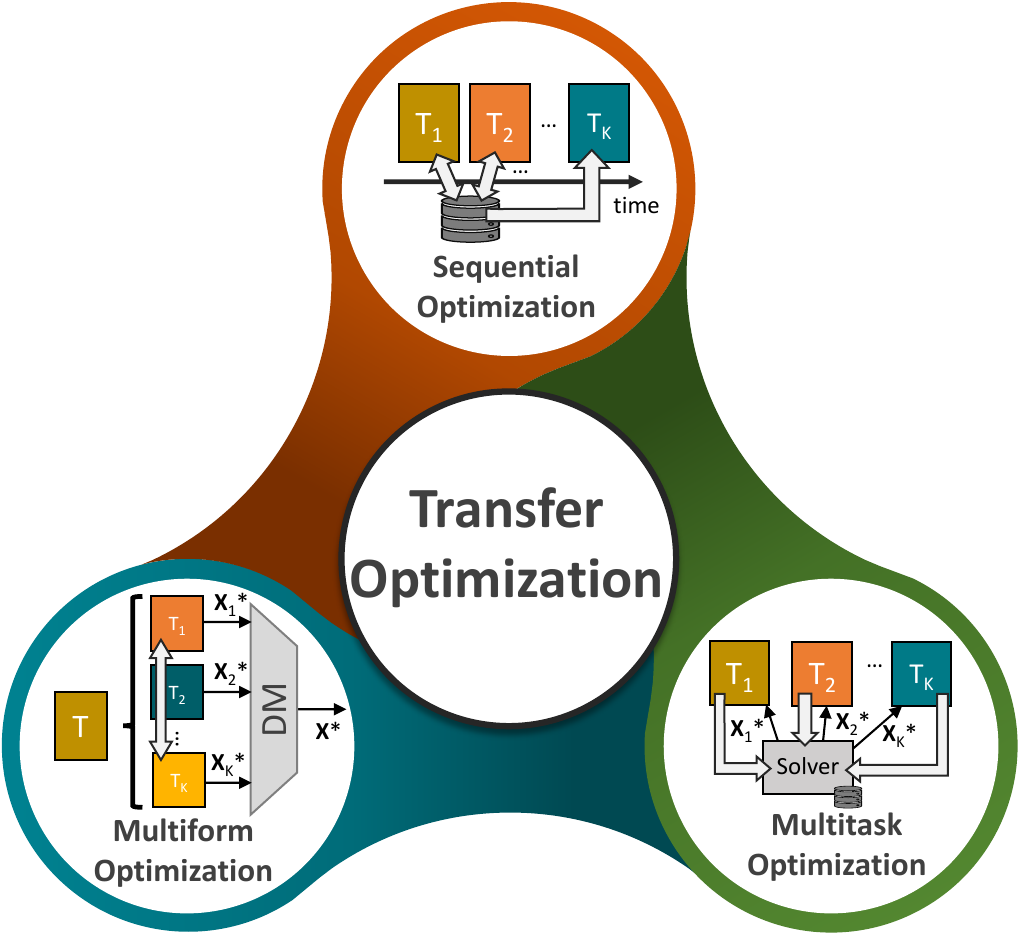}
	\caption{Different forms in which Transfer Optimization can be realized.}
	\label{fig:to}
\end{figure}

This trend has also permeated the optimization research area with the advent of the so-called Transfer Optimization paradigm \cite{gupta2017insights}, which has gathered significant attention from the community. Similar to its modeling counterparts, the \emph{raison d'\^etre} of this young research area is to leverage the knowledge acquired when solving one optimization problem to better address other problems, whether they are related or not. Embracing this overarching goal, three different paradigms have been identified in the landscape of Transfer Optimization: i) \textit{sequential transfer}, in which optimization tasks are solved sequentially (knowledge flows from one task to another once the former has been solved), ii) \textit{multitasking}, whose objective is to tackle several concurrent problems in a simultaneous fashion; and iii) \textit{multiform multitasking}, which addresses a single optimization problem by deriving alternative formulations and solving them simultaneously. {\color{black}Among these three categories, \textit{multitasking} is considered the one in the limelight of the research community, partly due to the capital role played by Evolutionary Computation in the development of renowned multitasking solvers}. This noted relevance has forged the above-mentioned term of Evolutionary Multitask Optimization, also regarded as Evolutionary Multitasking \cite{ong2016towards}. {\color{black}This paradigm refers to the adoption of concepts, operators and search strategies conceived within Evolutionary Computation for tackling multitask optimization scenarios. Figure \ref{fig:to} illustrates the relationship between all these forms of Transfer Optimization.}

In Evolutionary Multitasking, methods to realize knowledge transfer between the optimization tasks under consideration are central for the overall effectiveness of the multitasking algorithm itself. As a result, many works arising in this research area have striven for new knowledge transfer mechanisms, targeting to promote the exchange of information among related tasks and to minimize the potentially negative impact of transferring information among unrelated problems on the convergence of the search process. Although this motivation suffices for pursuing new research endeavors along this direction, there are more urgent needs in the field. These concerns can be described by fundamental questions (FQ) that lie at the very heart of the field itself:
\begin{itemize}[labelsep=3.5em, leftmargin=*]
	\item[{\parbox[t]{0.0\linewidth}{FQ1\\ \emph{Why?}}}] \parbox[t]{1\linewidth}{Does the Evolutionary Multitasking paradigm address a scenario that can be considered \emph{plausible} in practical settings? Does the simultaneous optimization of several related problems occur in real-world applications? Are there scenarios that can be approached using multitask optimization? In essence, is the main motivation of this research area justified by an informed evidence of its real-world applicability?}
	
	\item[{\parbox[t]{0.00\linewidth}{FQ2\\ \emph{What?}}}] \parbox[t]{1\linewidth}{Are evolutionary algorithms used for multitask optimization called by their preexisting names? Are we using the correct terminology and avoiding ambiguities? Are advances made in this area coherent with the state-of-the-art in meta-heuristic optimization, Evolutionary Computation and Swarm Intelligence?}
	
	\item[{\parbox[t]{0.00\linewidth}{FQ3\\ \emph{How?}}}] \parbox[t]{1\linewidth}{Is the performance of multitask optimization approaches measured fairly? Are benchmarks created on purpose for pairing problems with already known correlation properties? Should research studies account not only for the fitness improvements yielded by knowledge transfer, but also the implications of multitasking in terms of the computational effort required for the purpose? When solving real-world optimization problems, do we really obtain a profit by addressing them together via multitasking when compared to the case when the problems are solved in isolation from each other with competitive single-task optimization algorithms?}
\end{itemize}

After years of activity that have been summarized in recent surveys on Evolutionary Multitasking \cite{osaba2022survey,xu2021multi}, we firmly believe that it is the moment to expose and reflect these crucial concerns. Solid and informed answers to these fundamental questions are still lacking, which can lead to undesirable developments and outcomes of no practical value in the future of this field. To avoid them effectively, we herein spur an open constructive debate around the above issues, establishing the reasons why this discussion is of paramount importance for the evolution and practicality of the research area.

The rest of this manuscript is structured as follows: Section \ref{sec:what} describes briefly the basic concepts of Evolutionary Multitasking, as well as friction points that this paradigm maintains with other areas of Evolutionary Computation. Section \ref{sec:fund} elaborates on the three fundamental questions stated above. Finally, Section \ref{sec:prospective} ends this short communication by offering our prospects for the area, based on the conclusions drawn from the discussion held on the fundamental questions.

\section{Evolutionary Multitask Optimization: Concepts and Relationship to other Areas in Meta-heuristic Research} \label{sec:what}

As mentioned before, multitasking postulates that different yet concurrent optimization problems or tasks are simultaneously solved together. The goal is hence to obtain a good solution to every one of such problems, possibly exploiting as efficiently as possible the knowledge of every problem captured by the solver over the search \cite{ong2016evolutionary}. Mathematically, a multitasking environment comprises $K$ optimization tasks $\{T_k\}_{k=1}^K$ defined over as many search spaces $\{\Omega_k\}_{k=1}^K$. Without loss of generality, we assume that each task $T_k$ is a single-objective optimization problem driven by its own fitness function $f_k : \Omega_k \rightarrow \mathbb{R}$, where $\Omega_k$ is the search space over which the argument $\mathbf{x}$ of $f_k(\cdot)$ is defined. If we assume that all tasks should be minimized, the main goal of multitasking is to find a set of solutions $\{\mathbf{x}_1^{\ast},\dots,\mathbf{x}_{K}^{\ast}\}$ such that \smash{$\mathbf{x}_{k}^{\ast} = \arg \min_{\mathbf{x}\in\Omega_k} f_k(\mathbf{x})$}. 

Considering this formulation, two algorithmic strategies for tackling multitasking scenarios can be identified in the existing literature, which motivate the widespread adoption of meta-heuristic evolutionary and swarm intelligence solvers in the area: 
\begin{itemize}[leftmargin=*]
\item The execution of a single search procedure on a unique population $\mathbf{P}=\{\mathbf{x}^p\}_{p=1}^P$ of candidate solutions. Since all the solutions are contained in a single population, the challenge is to define a unified search space $\Omega_U$ over which candidates can be encoded, evolved, and decoded back to the specific search spaces $\Omega_k$ of every problem for their evaluation. Therefore, individuals evolve over $\Omega_U$, translating them to each independent search space $\Omega_k$ when required by means of an encoding/decoding function $\xi_k: \Omega_k\mapsto \Omega_U$. An evident benefit of this approach is the implementation of a single set of search operators, which gives rise a lower computational complexity of the search process. By contrast, it requires a proper design of the unified search space and encoding/decoding functions, which is not always straightforward to realize. This strategy is adopted by the family of multifactorial optimization methods, which resort to the concept of skill factors to drive the exchange of knowledge among tasks through the crossover operator and the unified search space. Among them, the influential Multifactorial Evolutionary Algorithm (MFEA) is without any doubt its most representative algorithm \cite{gupta2015multifactorial}.
	
\item The execution of different search processes running in parallel, one for each problem under consideration. In accordance with a knowledge sharing policy, information is exchanged between such search processes, either periodically or conditioned on the partial results of the search ({\color{black}e.g., a significant phenotypical change in the best solution of the task}). Under this strategy, each search procedure operates on a task-specific population of individuals $\mathbf{P}_k=\{\mathbf{x}_k^p\}_{p=1}^{P_k}$. In this case, each population runs over the search space $\Omega_k$ in which the task $T_k$ is defined. Regarding the sharing of genetic material, this is usually materialized through the exchange of individuals among different populations, involving this fact the existence of a mapping function $\Gamma_{k,k'}:\Omega_k\mapsto \Omega_{k'}$ for the translation of a solution $\mathbf{x}_{k}^p\in\Omega_k$ to the search space $\Omega_{k'}$ of task $T_{k'}$. In this second strategy the design of the search operators applied locally on each $\mathbf{P}_k$ is straightforward, as it only depends on the characteristics of the problem and its search space $\Omega_k$. {\color{black}By contrast, it generally scales worse with increasing $K$ than the previous single-population strategy, and requires defining a policy for the exchange of individuals among each pair of tasks.} This situation could increase even further the overall complexity of the multitasking solver. Devising problem-specific populations and search algorithms are the strategy adopted by multipopulation-based multitasking approaches.
\end{itemize}

For further information about how algorithmic schemes based on these two strategies work, we refer our readers to comprehensive surveys recently published in \cite{osaba2022survey,tan2021evolutionary,wei2021review,xu2021multi}. Among them, MFEA \cite{gupta2015multifactorial} and MFEA-II \cite{bali2019multifactorial} stand out as the arguably most influential works in the field.

\begin{figure}[h!]
\centering
\includegraphics[width=\columnwidth]{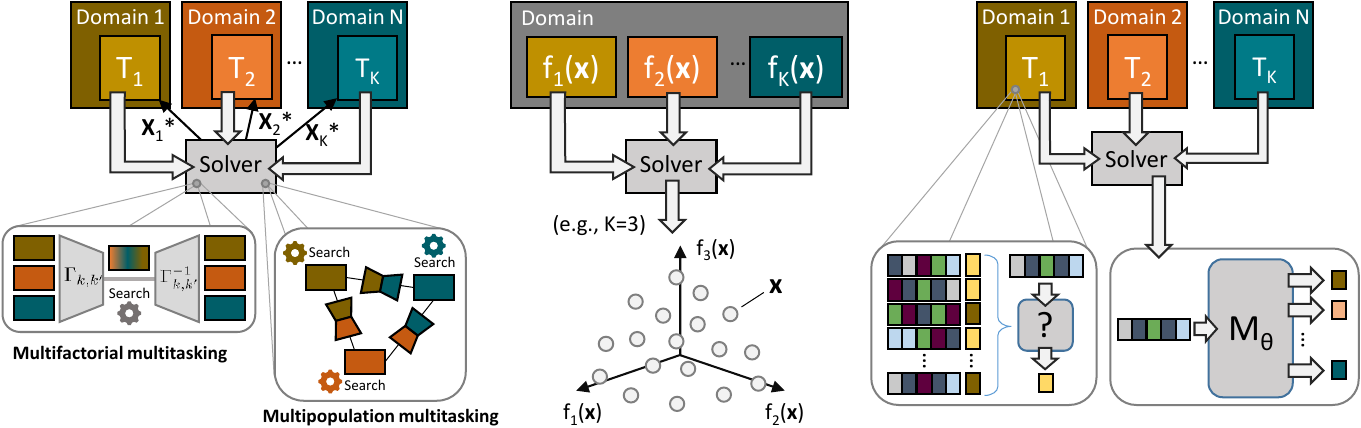}
\caption{Diagrams showing the differences between multifactorial optimization and multipopulation-based multitasking (left); multi-objective optimization (center); and multitask learning (right), the latter particularized for the case in which the solver aims to learn a model $M_\theta$ capable of addressing several supervised learning task at the same time.}
\label{fig:frictions}
\end{figure}

Figure \ref{fig:frictions} illustrates the strategies followed by multitasking-oriented algorithm described above. Based on these descriptions, the reasons for the marriage between Evolutionary Computation and Multitasking Optimization become clear and evident: {\color{black}populations serve as the knowledge base of the search algorithm, whereas the application of customized crossover operators (unified search space) or the exchange of individuals (multipopulation approaches) allow for the transfer of knowledge between optimization tasks}. Therefore, many research areas in Evolutionary Computation and Swarm Intelligence can be largely influential for the Evolutionary Multitasking paradigm. {\color{black}Key concepts such as co-evolution \cite{potter1994cooperative}, multi-population strategies \cite{ma2019multi}, archiving criteria \cite{laumanns2002archiving}, estimation of distribution algorithms \cite{hauschild2011introduction}, {\color{black}surrogate models \cite{forrester2009recent}}, parallel evolutionary algorithms \cite{cantu2001migration} and cellular genetic algorithms \cite{alba2009cellular} have been already considered by the community when proposing new multitasking solvers.} From a general standpoint, any research avenue in Evolutionary Computation that permits to split search resources can be thought to be capable of accommodating different tasks over the search process. Consequently, it can be extrapolated for multitasking setups by solely considering the existence of multiple problems in their inner working.  {\color{black}Thus, mechanisms such as sub-populations \cite{song2019multitasking}, sub-archives \cite{chen2019adaptive}, heuristics running in parallel or different neighborhood topologies (including cellular automata \cite{osaba2021mfcga}) are being actively investigated in the context of Evolutionary Multitasking}.

A further consequence of the above symbiosis is the controversy that often ignites around the similarity between multitasking optimization and multi-objective optimization, i.e., problems that comprise more than one objective function, all defined over the same search space and possibly conflicting with each other. {\color{black}The fact that multi-objective optimization considers several objectives and that such problems are often approached via Evolutionary Computation have propelled even further this controversy.} Certainly, conceptual overlaps exist among both research areas (such as the optimization of a group of objective functions), but both paradigms are different from each other. First, Evolutionary Multitasking aims to leverage the parallelism that brings a population of solutions in order to harness potential synergies between the problems at hand. Each of these tasks has its own solution space, often requiring the use of an encoding/decoding strategy for knowledge transfer. On the contrary, multi-objective optimization seeks a set of solutions that balance different conflicting objectives, but defined over an unique search space. {\color{black}Nevertheless, multi-objective problems can actually be tackled within a multitasking setup,} in which the tasks to be solved are multi-objective optimization problems \cite{gupta2016multiobjective}.

Other areas of friction can be found in the connection between multitask learning and multitask optimization, in the sense that we can construct data-based models capable of learning to solve several modeling tasks at the same time as a multitask optimization problem. To this end, solutions $\{\mathbf{x}_k^\ast\}_{k=1}^K$ sought by multitask optimization should represent the parameters of a model (as done, for instance, in symbolic regression with evolutionary programming). We can conceive multitask optimization as a possible alternative to deal with multitask learning problems, but it is not the only way to do it, nor is multitask learning the unique application for Evolutionary Multitasking. {\color{black}Differences and similarities between concepts and fields close to Evolutionary Multitasking are visually depicted in Figure \ref{fig:frictions}.}

\section{Fundamental Issues in Evolutionary Multitask Optimization} \label{sec:fund}

\begin{figure}[h!]
	\centering
	\includegraphics[width=0.9\textwidth]{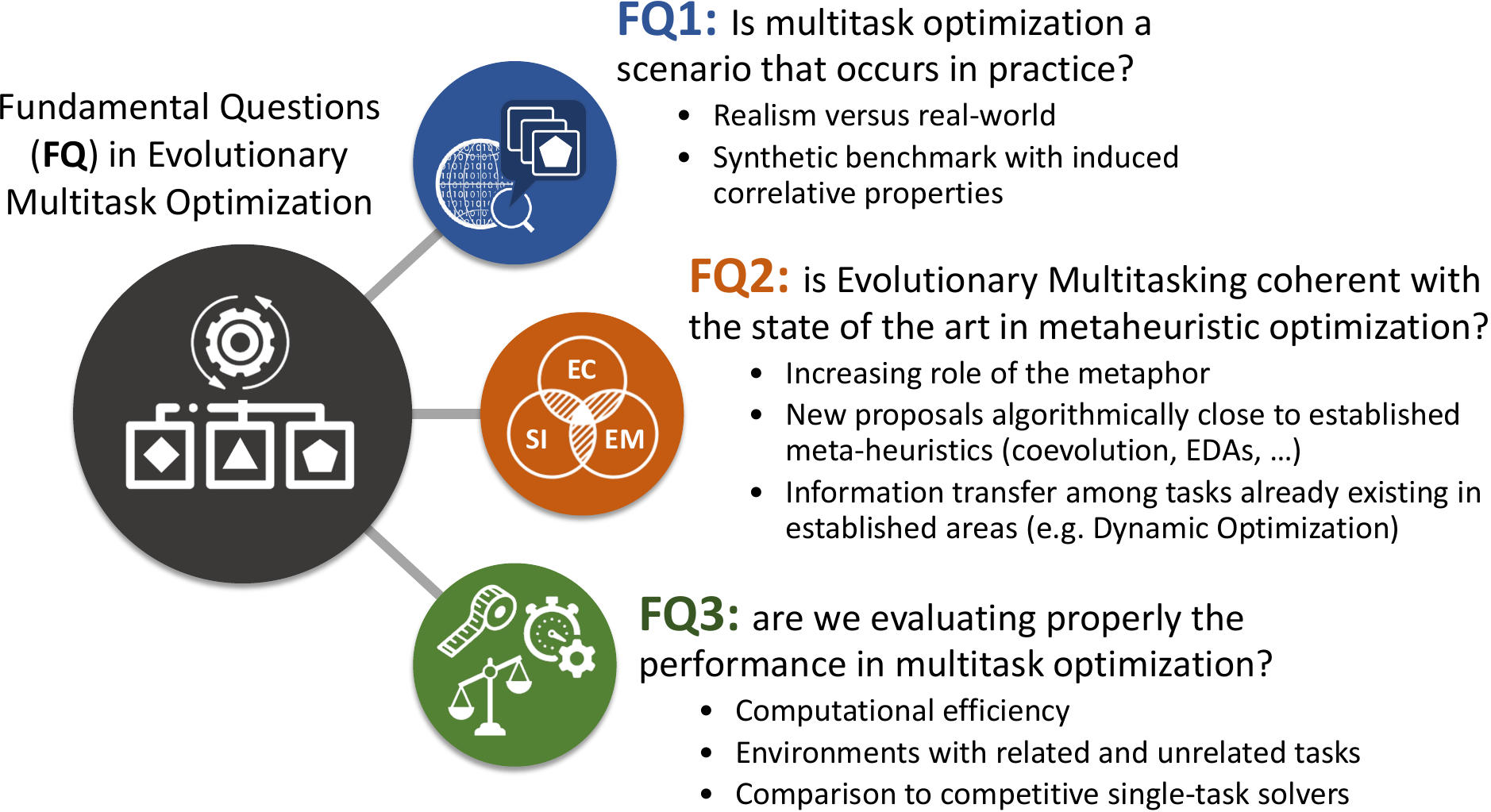}
	\caption{Schematic diagram showing the three fundamental questions (FQ) of Evolutionary Multitasking (EM) discussed in this paper. In the plot EC stands for \emph{Evolutionary Computation} and SI for \emph{Swarm Intelligence}.}
	\label{fig:issues}
\end{figure}

An upsurge of contributions has been noted around Evolutionary Multitasking and related concepts, as has been made clear in recent survey papers around the topic \cite{osaba2022survey,xu2021multi}. However, some fundamental questions need still to be clarified. In this section, we go deep into these main issues, exposing the main concerns that should be addressed by the community on each of these aspects: plausibility of the problem statement (Section \ref{sec:plausibility}), novelty of algorithmic advances in the field (Section \ref{sec:novelty}) and rigor in the evaluation methodology (Section \ref{sec:report}). Figure \ref{fig:issues} summarizes such concerns.

\subsection{FQ1 (Why?): Is Multitask Optimization a concept that occurs in the real-world?} \label{sec:plausibility}

The first issue that requires urgent attention from the whole multitasking community is the lack of convincing reasons why optimization problems should be solved in a simultaneous fashion. Although this claim may sound superficial, the whole paradigm relies on the idea of a temporal concurrence of optimization problems, i.e., multiple problems originate \emph{at the same instant of time}. Historically, the optimization field has focused on solving single problems under different circumstances: landscapes with challenging properties (e.g., ruggedness, multimodality), dynamic objective functions/constraints, {\color{black} or long fitness evaluation processes}). There have been specific cases when information about a problem has been reused for improving a new search process, either for the same problem (as in dynamic optimization, due to the non-stationary nature of its objectives and constraints) or for other related tasks (a change in the parameters defining the problem is usually tackled by feeding the population with previous solutions and restarting the search). In all such cases, tasks among which information has been transferred appear sequentially over time. Given this prior art, which suggests that in practice problems emerge sequentially, the research community should clarify, supported by evidence, whether independent optimization problems \emph{really} appear simultaneously in real-world environments. Even if this statement held, it should be verified whether such problems can be solved better by using multitasking solvers than in isolation with each other, by resorting to state-of-the-art single-problem solvers. FQ3 will later revolve on this second part of the issue.

When it comes to the real-world applicability of multitask optimization, we acknowledge the advance taken in this direction by an analysis recently published in \cite{gupta2022half}, where several real-world applications are described for multitask optimization. {\color{black} This work must be considered as a step towards assessing the applicability of the field to real-world optimization scenarios. Specifically, some of the real applications described in this work relate to 1) the use of Evolutionary Multitasking methods in data science pipelines, such as feature selection, hyper-parameter tuning, or neural architecture search; 2) the simultaneous planning of unmanned vehicle systems as a part of a wider robotic swarm; or 3) the concurrent optimization of manufacturing processes. Other efforts invested by the community for adapting the multitasking paradigm to real applications include last-mile logistics \cite{feng2020explicit} or the optimization of software testing  procedures \cite{sagarna2016concurrently}.

Despite the contribution of \cite{gupta2022half} to the applicability of multitask optimization,} we advocate for a step beyond these efforts: studies must not only report that multitasking \emph{can be applied} to real-world scenarios and problems, but also prove that multitasking \emph{is} utilized to solve real-world problems by virtue of its provided performance gains. Some fraction of the works reviewed in \cite{gupta2022half} cannot be considered to stem from a situation or a scenario encountered in a real use case. One of such examples is \cite{martinez2021adaptive}, which is referred in \cite{gupta2022half} as an example of real-world application related to \textit{evolving embodied intelligence}. In \cite{martinez2021adaptive} an evolutionary multitasking method is proposed for evolving multiple reinforcement learning agents coping with tasks of varying levels of complexity. The study demonstrates that transfer optimization can be a promising approach for solving such tasks, but also concludes that other multitask reinforcement learning algorithms yield closer levels of performance (success rate), but require a vastly lower computational complexity. In other words: \emph{feasibility} does not imply \emph{practical applicability} supported by superior performance and/or computational efficiency.

Further along this line, real-world optimization problems are known to be largely diverse in their search spaces, objective functions and/or constraints. Given this heterogeneity, can it be expected that concurrently originating real-world optimization problems will feature any complementarity or synergy when tackled simultaneously via multitasking in practical settings? The origins of this specific concern trace back to the inception of the multitasking research area itself. Almost all areas of research in the optimization field depart from a need or problem observed in a real-world setup. Multitasking, however, started from conjectural premises about the similarity between the solutions to different problems. Symptomatically, application fields such as logistics, medicine or energy have afforded several real-world benchmarks for modeling and optimization. {\color{black}On the contrary, evolutionary multitask optimization has so far dealt with synthetic benchmarks comprising functions that are endowed with synergies that offer synthetic advantages for multitask optimization.} That is, the research has gone \textit{from the technique to the synthetic problem creation} rather than \textit{from the real-world problem to the technique}, which ensures that scientific advances naturally flow towards practical impact. For academic purposes, insights can be surely drawn by hypothesizing on the problem based on intuition and common reasoning. But conversely, the lack of supporting evidence for the formulated hypotheses hinders and puts to question whether multitasking effectively contributes to the ultimate objective of optimization research: to solve real-world problems as efficiently as possible \cite{osaba2021tutorial}. Naturally, we should solve optimization problems as soon as they arise in the real world, rather than waiting for some time until several tasks are collected expecting some exploitable synergies among them (which brings to question how long one can and should wait for, or how long the first problem can be delayed in the process).

On the positive side, some recent works have presented preliminary results about the possible use of Evolutionary Multitasking as an effective tool to improve the search efficiency in computationally expensive problems. This is the case of \cite{wang2021surrogate}, which showcases the adoption of surrogate-assisted MFEA to efficiently address robust minimax optimization problems, which also seek the worst scenario in which the problem can be formulated. This aligns with other works on the suitability of Evolutionary Multitasking to tackle problems characterized by time-consuming function evaluations and/or diverse problem formulations \cite{liao2020multi,gupta2015evolutionary,rios2021multi}. Among them it is worth noting the multitask variant of Bayesian optimization \cite{swersky2013multi}, which allows for transferring knowledge between hyper-parameter tuning tasks (a computationally burdensome problem on its own) of machine learning models learned over different albeit related domains. {\color{black}Such works delineate an interesting road of opportunity for Evolutionary Multitasking, but more principled and fair comparisons to state-of-the-art algorithms are still needed, {\color{black}together with clearer evidence of the real-world applicability of Evolutionary Multitasking.} In references as the ones cited above, comparisons are often done against non-multitasking versions of the same solver to account for the benefits yielded by exploiting the relationships between tasks, but override any comparison to alternative approaches known to perform better and/or more efficiently when addressing the considered tasks independently.} 

This issue also raises another concern when it comes to competitions and benchmarks: are multitasking setups for benchmarking purposes created with highly correlated problems to show that algorithms perform well by exploiting such correlations? In contributions related to evolutionary multitasking it is common to have the proposals validated over benchmarks comprising synthetic functions, which are used for evaluating the quality of the developed methods. These benchmarks combine correlated problem instances that create favorable scenarios for multitasking approaches. For instance, the benchmark problems defined for {\color{black}the 2022 edition of the IEEE Congress on Evolutionary Computation} (CEC'2022) competition on evolutionary multitasking for single-objective continuous optimization \cite{da2017evolutionary} defines several scenarios comprising tasks with varying levels of intersection and similarity between their global optima in the unified search space. Such tasks are essentially rotated and/or shifted versions of well-known test functions for continuous optimization, whose rotation matrices and shift vectors are tailored to induce different properties in their search spaces. The benchmark itself includes a scenario with \emph{No Intersection and Low Similarity} tasks so as to account for the robustness of Evolutionary Multitasking methods under unfavorable conditions. Even if allowing for a proper comparison, the fact that such benchmark scenarios are constructed by rotating and shifting synthetic test instances keeps into question whether the correlation properties enforced in the scenarios are representative of a real-world situation.

Prospective authors should provide rationale whether the simultaneity of real-world problems in any given practical scenario 1) allows for the existence of synergies between tasks that makes exploitation worthwhile; and 2) that approaching such problems with a multitasking solver is compliant with the non-functional requirements of the scenario ({\color{black}e.g., would any driver be willing to share the source/destination node of his/her route with a remote machine in the cloud?} Is the sharing of industrial production schedules across different plants a reasonable premise?). For these reasons, we should avoid independent/separate comparison of methods using tailored benchmark problems without any connection to the limitations of real-world scenarios. {\color{black}And most importantly, the overall assumption of correlated global optima among multiple problems that lies at the core of multitask optimization should be proven to hold in practice with irrefutable evidence. Only when this important factor is verified multitask optimization methods can be designed, performing consistently with respect to the practical constraints and properties of their targeted scenario.}

This lack of applicability can be also noted in the forums where works related to Evolutionary Multitasking are currently published. It is often the case that new contributions springing in this area are reported in high-quality journals and tier-one conferences dedicated to Artificial Intelligence, with emphasis on algorithm developments in Evolutionary Computation and Swarm Intelligence. On the contrary, contributions dealing with evolutionary multitask optimization are rarely published in conferences and journals specialized in a certain application field. This can be a sign of the questionable applicability of the field since its inception in 2015, since these latter scientific venues are more concerned with the applicability of the technical proposal in real-world settings. Contributions are more strictly evaluated in terms of the plausibility of the scenarios and the gains/suitability of the proposed solution rather than in terms of its algorithmic sophistication. Prospective real-world applications of evolutionary multitasking for readerships that are experts on such applications will have a significantly higher soundness than more algorithmic proposals evaluated over favorably created synthetic benchmarks.

On a positive note, we eagerly encourage authors to embrace the use of previously gathered knowledge when facing new real-world optimization tasks, even if this exploitable knowledge is generated by solving synthetic problems that may provide useful information for the optimization process. {\color{black}In any case, we call for a deep reflection around the benefits of solving real-world problems in a simultaneous fashion, neglecting any prior knowledge and assuming a \emph{cold start} for the overall process}. A more reasonable paradigm in practice is to consider huge archives of previously gathered knowledge (solutions to other problems) before solving a new problem, discover which previous problems are most similar to the one to be solved (e.g., by virtue of meta-features extracted from each problem), and use stored solutions as a starting point for addressing the new task (for instance, by seeding the initial population). This specific type of sequential knowledge transfer can be more impactful in real-world applications{\color{black}, whenever contributions to realize this paradigm give credit to advances achieved over the past in related areas of metaheuristic research, including archive memories or landscape analysis, to mention a few.}

{\color{black} To conclude our elaborations around FQ1, it is worth pausing at some realistic application scenarios tackled with transfer optimization algorithms in recent studies, which can potentially benefit from effective transfer optimization mechanisms: for instance, planning the routes of several vehicles in a city, or the design of simulation-based processes. Unfortunately, scenarios as such often neglect practical constraints that could hinder the exchange of knowledge among the involved tasks: for instance, are candidate solutions to a given task shareable by all means with third-parties due to privacy/confidentiality restrictions? Can one expect that in e.g. logistic planning, {\color{black}information about the routes followed by a fleet of vehicles can be exchanged between different companies/travelers?} What if the genotype of exchanged solutions can reveal sensitive information about the problem, as it could occur in the therapeutic prescription in medical treatments, or production scheduling of manufacturing plants? What about the existence of conflicting goals and/or precedence constraints among tasks? Most setups currently addressed via multitask optimization oversee these important aspects (e.g., correlation of the problems, number of decision variables, concurrency in the creation of problems) that can make the exchange of knowledge between tasks unreliable/non-robust in practical settings, even if it is proven to be algorithmically beneficial in ideal conditions. State-of-the-art solvers designed for single problems, however, can perform competitively disregarding whether these aspects hold for the scenario at hand.}

\subsection{FQ2 (What?): Are evolutionary algorithms used for multitask optimization coherent with the state-of-the-art in meta-heuristic research?}\label{sec:novelty}

If we compare Evolutionary Multitasking to other areas in optimization research, it is undeniable that it still remains in an early stage of development, with a long road ahead with plenty of challenges and discoveries. Despite its infancy, this area is {\color{black}timidly} approaching similar practices to the controversial ones observed in Evolutionary Computation and Swarm Intelligence: the explosion of methods claiming to portray a set of own algorithmic peculiarities, but very similar in their essence to other solvers that have prevailed in the field for decades \cite{sorensen2015metaheuristics,del2019bio}. {\color{black}This trend has led to hundreds of meta-heuristic approaches with scarce differences at their core,} both among them and with respect to traditional optimization heuristics \cite{molina2020comprehensive}. 

{\color{black}Having said that, we have noted that, until now, the number of metaphor-based algorithmic proposals is much lower in Evolutionary Multitasking than in other areas of optimization research. The \emph{algorithmic essence} of every new Evolutionary Multitasking proposal should concentrate on novel strategies to allocate resources among tasks and/or the way knowledge is modeled and transferred among tasks. By all means, they should leave aside the evaluation of sophisticated-yet-already-existing search operators that do not account for the correlation between tasks whatsoever. For this reason, authors should work on properly highlighting the \emph{true contribution} of their proposed method to the multitasking community. This is the case in most studies to date, but the latent risk of witnessing an uncontrolled production of nature-inspired Evolutionary Multitasking method calls for the adoption of measures to avoid it effectively. All future algorithms \emph{must} highlight their uniqueness and similarity to prior works, dissecting each of their algorithmic steps without any metaphorical reference, preferably in the context of a concrete real-world application.}

In connection to this issue, we have also noted name-change research trends, i.e., some algorithmic strategies for Evolutionary Multitasking, when inspected closely, can be found to be refactored versions of already existing concepts in meta-heuristic optimization research. Archive memory, multiple populations, estimation of distributions along the search, topologies, or migration strategies have been considered in different contributions over the short history of Evolutionary Multitasking, and should continue stimulating new developments in the area. {\color{black}But prospective studies using them should avoid any refurbishing of their names, and should clearly and explicitly analyze whether new proposals skim any established area in Evolutionary Computation or Swarm Intelligence.} {\color{black}Undeniably, such studies provide fresh perspectives on previous population-based optimization methods. Nevertheless, their connection and incremental nature with respect to the state-of-the-art should be highlighted rather than concealed under a refactored presentation.}

This name changing trend affects beyond certain multitasking approaches proposed to date, permeating into the roots of the Transfer Optimization paradigm. As anticipated in the introduction, three different paradigms are identified in Transfer Optimization: {\color{black}sequential transfer, multitasking optimization, and multiform multitasking.} Although the last two paradigms introduce new concepts and research pointers, sequential transfer has been extensively studied in the literature along the years. Reusing what has been learned in the past for expediting the search process when optimization a new task (or a new version of it) is a well-known strategy in dynamic optimization \cite{nguyen2012evolutionary}, particularly in the case with recurring change patterns. In most schemes departing from the latter casuistry, the solver must store and {\color{black}retrieve from stored solutions encountered in the past (\emph{explicit memory})}, injecting them into the population to better accommodate changes in the problem formulation and/or constraints \cite{Yazdani2021A}. 

This design principle clearly overlaps what is pursued in sequential transfer optimization, unless it is assumed that the variation undergone by the problem is abrupt enough for the variation to be considered a \emph{new related problem}. In this case one can argue that the transfer of knowledge goes beyond storing information of the current problem and retrieving it for solving a variation of the same task. If the problem at hand varies enough to be regarded as a completely new optimization task, the challenge to exploit knowledge captured prior to the changes becomes more complex, making sequential transfer optimization a research area on its own. In this case, differences between the new problem formulation and its preceding one ({\color{black}e.g., heterogeneous search spaces}) may impose search strategies and knowledge modeling strategies (correspondingly, learnable solution representations or subspace alignments) that step further the current state-of-the-art in dynamic optimization. In any case, advances in this direction should acknowledge their links to dynamic optimization, even if extending what has been done so far in this area.

Summarizing, in FQ2, we underscore the fact that many methods proposed in the context of Evolutionary Multitasking and Transfer Optimization can -- and must -- be regarded as extensions of already published methods in meta-heuristic optimization research. This is noteworthy in dynamic optimization, where several approaches for information transfer among time-varying problems (essentially, to reuse past solutions) have been proposed in the recent past \cite{branke2012evolutionary}. Such variants have been also used in the context of Transfer Optimization and Evolutionary Multitasking ({\color{black}e.g., archive memories in \cite{chen2019adaptive} or surrogate modeling assisted information transfer in \cite{9377470}}). Unfortunately, in several cases these variants come along with a change of terminology, which does not fairly establish links to other areas of optimization research. We firmly advocate for studies in which the connections of newly proposed algorithms to traditional areas of optimization research are identified, {\color{black}so that ambiguities are minimized, and meaningful advances are achieved.} In this regard, a remarkable attempt is made in the survey published in \cite{xu2021multi} to describe the theoretical foundations of evolutionary multitask optimization, examining its mathematical ingredients and clearly defining each mechanism and solving strategy. More along this line, in \cite{osaba2022survey}, the main concepts of multitasking optimization are differentiated from other potentially colliding fields, such as multi-objective optimization or transfer learning. However, more fine-grained clarifications are still in need: new multitask optimization methods to appear in the future should include a thorough examination of the similarities and differences of their algorithmic components to established methodologies in other areas of meta-heuristic optimization research.

\subsection{FQ3 (How?): Are evolutionary multitasking approaches evaluated properly, with right metrics and in fair comparison benchmarks?}
\label{sec:report}

The last issue we bring to discussion in this short communication relates to the practices adopted for evaluating the performance of Evolutionary Multitasking solvers. To begin with, a bad practice observed in studies related to this area is to conduct experiment by comparing new multitasking solvers to other multitasking methods, mostly in terms of fitness value (or any other indicator of the quality of the solutions encountered for the tasks under consideration). This comparison is needed for verifying that the proposed solver attains gains with respect to the state-of-the-art in Evolutionary Multitasking. For the sake of a fair experimental analysis, discussions held around the results should clarify, in an informed fashion, whether the gains can be attributed to better search operators, more effective knowledge transfer mechanisms, a more fine-grained parameter setting, or any other aspect of the multitasking solver. But beyond all, experiments should be completed with a rigorous and mandatory comparison of the evolutionary multitasking solver to the alternative case where tasks are solved separately from each other, using competitive state-of-the-art algorithms, and subject to a similar computational effort budget. 

{\color{black} A work embracing this recommendation is the recent work presented in \cite{osaba2022multifactorial}, where a cellular evolutionary multitasking approach is proposed to concurrently solve multimodal optimization problems. The experimentation comprises 14 different instances drawn from the CEC’2013 competition on multimodal optimization, yielding several valuable insights in connection to FQ3. To begin with, experiments prove empirically that knowledge sharing among multimodal optimization tasks is beneficial with respect to individual problem solving with the same solver if the global optima of the problems at hand are related to each other. This assumption falls far from real-world multimodal problems, which usually have a very limited number of global solutions and a higher quantity of local solutions: it is more likely in practice that a global solution of one problem becomes aligned with the local solution of another problem, leading to a potentially adverse exchange of knowledge if a multitask solver is adopted.

Secondly, the performance of the proposed cellular evolutionary multitasking solver is compared to that of other multitasking algorithms, rendering a superior performance. However, and most importantly, the performance of all multitasking methods considered in the study is compared against single-objective optimization algorithms suited for multimodal optimization that have ranked high in competitions on multimodal optimization held so far. This last experiment reveals that multitasking methods fail to perform competitively when they are compared to approaches designed specifically for multimodal optimization. In conclusion: advances in evolutionary multitasking are of no practical impact if the exploitation of the synergies among tasks is outgained by more efficient and outperforming single-task solvers. Fair comparisons must be made against representative solvers from the state-of-the-art of the optimization problems tackled by the multitasking method.} 

Furthermore, experiments held over multitasking environments should consider quantitative measures beyond fitness value statistics. One of the reasons to opt for a joint optimization of several tasks at the same time is to reduce the computational effort of the search for their solutions. This implies either a lower number of function evaluations needed for discovering solutions of a given level of quality or, alternatively, better solutions found under the same computational budget. This requires a major rethink on how results in Evolutionary Multitasking should be normalized and interpreted considering the amount of resources (memory, function evaluations) consumed over the search, not only in what refers to function evaluations, but also regarding the additional computation burden imposed by estimating, transferring and exploiting knowledge among tasks. {\color{black} Several preliminary ideas can be outlined for this purpose, including a normalization of the results with respect to the best solutions achieved by non-multitasking solvers for the problems at hand, repeatedly randomizing the experimental cases for them to comprise different mixtures of related and unrelated tasks, or reporting on performance metrics that account for the overall number of function evaluations inside the multitask solver. All in all, the community needs to differentiate between comparison studies among different multitask optimization approaches and between the multitask optimization proposal of the work at hand and competitive single-task solvers, deriving quantitative and fair measures of performance (that account for both optimality, convergence, and complexity) for both types of comparison studies.}

Finally, another concern with experimental methodologies followed to date by the Evolutionary Multitasking community is the examination of the performance of the developed multitask solvers in environments composed by unrelated tasks. In accordance with our reflections in Section \ref{sec:plausibility}, a large fraction of the works emerging lately in this area still conduct experiments over benchmarks composed by synthetic functions, in which synergies among tasks are forced to favor its exploitation during the multitask search. Besides the handcrafted nature of these benchmarks (which we already discussed in FQ1), experiments should be augmented with Evolutionary Multitasking scenarios composed by unrelated and stochastically generated tasks, so that the discussion is fed with performance results of the method when facing favorable and unfavorable scenarios given the purpose the multitask solver was designed for. 

An exception to this statement is the aforementioned CEC'2022 benchmark on evolutionary multitasking for single-objective continuous optimization \cite{da2017evolutionary}. In this benchmark one of the scenarios comprises tasks with no intersection between their global optima in the unified space, nor are their optimal solutions similar to each other. Interestingly, baseline results reported in \cite{da2017evolutionary} reveal that, when dealing with the scenario consisting of totally unrelated tasks, the performance of evolutionary multitasking (as represented by MFEA) degrades with respect to a naive single-objective, non-multitasking evolutionary solver that is actually not competitive in the tasks included in the scenario. {\color{black} The work on community detection over graphs using coevolutionary multitask optimization reported in \cite{osabay2020coevolutionary} also considers multitasking scenarios composed by fully correlated and uncorrelated problem instances. Furthermore, the performance of the proposed coevolutionary multitask solver is compared to single-objective, non-multitasking versions of the same solver. Conclusions drawn from this study shown that when tasks are related, the multitasking approach emerges as the best alternative, while no difference is perceived among the results of the coevolutionary multitask solver with respect to solving the tasks in isolation when they are unrelated to each other. 

Nevertheless, the work in \cite{osabay2020coevolutionary} exemplifies the early stage of development in which Evolutionary Multitask Optimization prevails: no comparisons to other established community detection algorithms ({\color{black}e.g., Louvain}) were made, even though they have been proven to perform competitively with respect to meta-heuristic algorithms used for this task. Furthermore, the experimental setup consisted of network instances of the same size ({\color{black}despite the important implications in terms of solution encoding and knowledge sharing those differences in genotypical length could entail}), and only the quality of solutions was examined in the discussed comparisons. Studies like \cite{osabay2020coevolutionary} can be largely improved if our reflections offered in FQ3 are followed.}

For this reason, until there is an algorithm proven to effectively pair related problems {\color{black}to be solved collectively, effectively, and universally, a large number of experiments} comprising related and unrelated problems selected fully at random should be conducted, including a mandatory comparison against solving each optimization task by competitive state-of-the-art methods. Should such a pairing approach eventually exists, the computational resources required to pair up related problems from a large pool of problems should be considered as well. Above all, we again emphasize that in real-world scenarios, optimization tasks with synergies and commonalities rarely occur (let alone arising simultaneously). For this reason, the degradation of the results in environments with unrelated tasks should be thoroughly evaluated, as it is the circumstance that the algorithm will most frequently encounter in practice. 

\section{Prospective: Something Else is Needed in Evolutionary Multitask Optimization}
\label{sec:prospective}

Transfer Optimization and Evolutionary Multitasking are at their dawn. A growing corpus of contributions are published on a daily basis, in top conferences and reputed journals \cite{xu2021multi,tan2021evolutionary,wei2021review}. This area has garnered much interest from the community working on optimization, blowing a fresh breeze of new developments and research directions over the field. 

Unfortunately, not all that glitters is gold: {\color{black}Evolutionary Multitasking needs urgent, sincere, and reflexive thoughts about fundamental questions that should be addressed for the prosperity of the area as a whole.} Solid grounds are still missing in terms of i) {\color{black} the practical applicability and competitiveness} of the multitasking paradigm; ii) the novelty and reciprocity of algorithmic proposal with respect to the state-of-the-art in optimization research; and iii) the fairness and rigor of the methodologies for performance assessment and comparison used to date. As a result, a growing strand of literature blossoms every day without convincing responses to the aforementioned issues. On a prescriptive note, we advocate for several specific actions that could bring light to the area:
\begin{itemize}[leftmargin=*]
\item[\rlap{\raisebox{0.3ex}{\hspace{0.4ex}\tiny \ding{52}}}$\square$] Informed evidence of the practical relevance of the multitasking paradigm should be given ({\color{black}e.g., by providing examples of real-world setups where the tasks at hand could co-occur,} together with the reasons why they might occur at the same time and why they could feature synergistic relationships with each other). {\color{black}Rationale offered in this regard should also specify the computational aspect favored by the adoption of multitask optimization, be it a faster convergence or a higher quality of the solutions discovered for each problem under consideration.}

\item[\rlap{\raisebox{0.3ex}{\hspace{0.4ex}\tiny \ding{52}}}$\square$] More studies showcasing the usage of evolutionary multitasking in real-world (not only \emph{realistic}) applications published in fora and journals specialized in the field at hand, {\color{black}leading to competitive levels of performance and efficiency with respect to state-of-the-art single-problem solvers}. {\color{black}Such studies should conform to good methodological practices to design, evaluate and implement meta-heuristic algorithms in real-world optimization scenarios \cite{osaba2021tutorial}.}

\item[\rlap{\raisebox{0.3ex}{\hspace{0.4ex}\tiny \ding{52}}}$\square$] Prospective works should consider and satisfy all specifications, requirements and constraints that one could encounter in a multitasking setup, especially when it comes to information sharing across tasks {\color{black}({\color{black}e.g., the confidentiality of solutions when shared to third-parties, or the obsolescence of the solution in time-varying tasks).}}

\item[\rlap{\raisebox{0.3ex}{\hspace{0.4ex}\tiny \ding{52}}}$\square$] Algorithmic contributions should focus on improvements that can be attributed to the strategy for knowledge transfer and exploitation, rather than to sophisticated/metaphor-based search operators that do not relate at all to the presence of multiple, potentially correlated optimization tasks.

\item[\rlap{\raisebox{0.3ex}{\hspace{0.4ex}\tiny \ding{52}}}$\square$] Every newly proposed algorithm should be complemented with a thorough analysis of possible algorithmic overlaps with other preexisting areas of Evolutionary Computation and Swarm Intelligence, thereby avoiding name-change research.

\item[\rlap{\raisebox{0.3ex}{\hspace{0.4ex}\tiny \ding{52}}}$\square$] {\color{black}The added computational complexity} of evolutionary multitasking methods should be regarded as a mandatory aspect for the evaluation of new proposals.

\item[\rlap{\raisebox{0.3ex}{\hspace{0.4ex}\tiny \ding{52}}}$\square$] Comparisons must be done to other multitasking alternatives, but also to competitive single-task solvers from the state-of-the-art subject to the same computational budget. {\color{black}The fitness of solutions delivered by multitask optimization should be normalized by that of the best-known solutions achieved by single-task solvers for the considered problems, so that the search efficiency of multitask solvers is gauged with respect to the state-of-the-art performance known for such problems.}
\end{itemize}

In no way the intention of this short critique is to condemn the achievements reached in Evolutionary Multitasking to date, nor do we intend to divert efforts away from this research area. Diametrically, we believe that Transfer Optimization and Evolutionary Multitasking deserve proper research attention. However, efforts to be invested in the future must guarantee that the right questions are addressed to dispel the doubts and concerns exposed in this manuscript. Otherwise, the field will maintain an uncontrolled growth without {\color{black}well-grounded conceptual, methodological, and practical rationale, eventually leading to a manifold of studies with little practical relevance.}

A global understanding and assumption of these needs by the community requires an explicit \emph{manifesto} that serves as a referential point of consensus. This is indeed the ultimate purpose of this open letter. Evolutionary Multitasking is still a young area facing a long road ahead to develop itself and showcase its postulated benefits in real-world applications, competing fairly against state-of-the-art single-instance {\color{black}solvers}. It is now the time to ensure that this road can be driven safely, and sure of reaching a meaningful destination.

\section*{Acknowledgements}

The authors would like to thank the Basque Government for its funding support through the ELKARTEK program and the consolidated research group MATHMODE (ref. IT1456-22).

\bibliography{biblio}
\end{document}